# Non-photorealistic image processing: an Impressionist rendering

**Amelia Carolina Sparavigna[1] and Roberto Marazzato[2]**
1 Department of Physics, Politecnico di Torino
2 Department of Automation and Computer Science, Politecnico di Torino
C.so Duca degli Abruzzi 24, Torino, Italy, I-10129, Italy

**Abstract**. The paper describes an image processing for a non-photorealistic rendering. The algorithm is based on a random choice of a set of pixels from those of the original image and substitution of them with colour spots. An iterative procedure is applied to cover, at a desired level, the canvas. The resulting effect mimics the impressionist painting and Pointillism.

## 1. Introduction

Traditionally, computer graphics pursued the reproduction of real world. Consequently, many efforts were devoted to the photorealistic approach of rendering and processing images. These researches achieved so impressing results such as those that we can admire in 3D animation films.
Photorealistic processing algorithms are fundamental in those processing required for scientific and forensic imagery, where they can remove noise or enhance object outlines in the image scene. However, it is not obvious that a photorealistic processing is always to be preferred, because it is not the only and best vehicle for communicate information. Let us consider for instance the hand drawing, which is clearly a non-photographic imagery. Hand-drawn illustrations can better explain a scene than photographic plates, because in illustrating complex phenomena, they can omit unnecessary details and propose only fundamental objects.
Non-photorealistic rendering focuses on to the development of algorithms for generating or processing images that embody the following qualities: emphasis of selected features, suppression of unimportant details, and use of stylization to suggest emotional structures. Therefore, a wide variety of expressive styles exists (1-4). Some algorithms are devoted to produce pseudo hand-drawing images (5). Other techniques are working with algorithms for visualizing vector fields (6), mainly based on the line integral convolution (LICs) (7,8).
Non-photorealistic rendering styles, inspired by artistic methods such as painting and drawing, but also technical illustration and cartoons, have been developed for digital art. We have then many rendering techniques simulating the painter's medium, with methods emulating the diffusion of ink through different kinds of paper, and of pigments through water for simulation of watercolour or oil painting. After a short review of Impressionist and Neo-Impressionist art, we discuss and propose some image processing methods to obtain an Impressionist rendering of images.

## 2. Impressionist and Neo-Impressionist art

Impressionism was a 19th-century art movement that began as an association of artists whose independent exhibitions brought them to prominence in the last quarter of the century. The name of the movement is derived from the title of a Claude Monet work. Characteristics of Impressionist paintings include visible brush strokes, emphasis on light and shadows, as changing qualities, achieved by accentuating the effects of the passage of time, the inclusion of movement as a crucial element of human perception and unusual visual angles (9). For Impressionist painters, images must be able to create a feeling and then it is not necessary an almost perfect reproduction of reality.
At the beginning of impressionist movement, photography was gaining popularity, inspiring the artist to seize the moment, in landscapes with their fleeting lights, or in the every-day life of common people. In painting realistic scenes, they emphasized vivid overall effects rather than details. In fact, Impressionists took advantage of the mid-century introduction of premixed paints in lead tubes, which allowed artists to work spontaneously, both outdoors and indoors. Previously, painters made their own paint media individually, by preparing and mixing pigment powders with linseed oil (10).

In Neo-Impressionist art, colours are applied side-by-side with as little mixing as possible, the optical mixing occurring in the eye of the viewer. The French art critic coined the name of this technique, Neo-impressionism, in 1886 to describe the painting by Georges Seurat.

Seurat is credited as the creator of Pointillism painting technique, in which dots of almost pure colour are juxtaposed on the canvas. The dots blend together in the observer's eye to create tones when one looks at the painting from a distance. The Italian equivalent, also developed from the studies being done in optics, is known as Divisionism. One of the main differences is that Divisionists used longer brush-strokes, rather than small dots of paint. Another difference is the type of subjects: Divisionists preferred the social themes.

Neo-Impressionist artists were influenced, without any doubt, by some science writers that published treatises on colour, optics and perception. One of them, Michel Eugène Chevreul, a French chemist, had perhaps the most important role on artists (11). In 1861, Chevreul, during the restoration of old tapestries, noticed that the only way to properly restore the missing wool was to take into account the influence of coloured yarns around it. He discovered that two colours juxtaposed, slightly overlapping or very close together, would have the effect of another colour when seen from a distance. Chevreul also realized that the halo that one sees after looking at a coloured gleam is due to retinal persistence of complementary colour. Neo-impressionist painters made extensive use of juxtaposition and complementary colours in their paintings.

**3. Impressionist rendering**

As discussed in (12), the painted impressionist technique produces scale-based effects. Far-field vision is realistic while near-field vision is confused, as summarized by D. Laidlaw (13). A few inches away from the canvas, it is possible to see how the strokes of the brush placed and mixed colours. This Impressionist effect has been the goal of many works from the beginning of non-photorealistic rendering. Through colouring the image as if it were produced by brush strokes, it is possible to have an accurate medium range Impressionist effect, with an image that is blurred, but with the general shapes of objects still recognizable. Hertzmann added multiple sizes brushes, the smallest ones being used to enhance details (14,15).

Impressionist renderings are proposed by Photoshop and GIMP image processing applications (16). The last program has the GIMPressionist tool where it is possible to choose paintbrushes and canvases, and adjust the orientation of brush and the flow of colour through the rendered painting.

As previously discussed, Pointillism is a style of painting in which small distinct dots of colour create the impression of a wide selection of other colours and blending. The depicted imagery emerges from disparate points. The technique then relies on the perceptive ability of eyes and mind of the viewer to mix the colour spots into a fuller range of tones. Used colours were usually limited in number: a half-tone technique can be pursued to have a Pointillist rendering. In Ref.17 such a Pointillist rendering was proposed, which, as in half-toning procedures, uses a limited palette of colours. A Voronoi scheme determines the random position of dots.

Here we propose a simple procedure to create an Impressionist, Pointillism-like rendering, starting from an input map. First, we choose the background of output image, which can be white or in agreement with the hue of the original image, or also the original image itself. Then we randomly choose those pixels of the original image, which have to be substituted by a colour spot.

For this choice, a random number $s \in \mathbb{N}$ is generated, between two values, $s_{min}$ and $s_{max}$. $s$ is the increment of two the loop counters, $p_i$ and $p_j$, ranging from 1 to $N_x$ and $N_y$, which span the rectangular image dimensions. In this way, we obtain an array of image pixels, at given positions $(p_i, p_j)$. Then the RGB colour tones for such pixels are determined: let us call the function describing colour tones $b(p_i, p_j, c)$, where $c$ is the index R,G and B of the colour tones. Since this array of pixel positions $(p_i, p_j)$ is regularly spaced, we further generate, for each pixel of the array, another couple of random numbers $\delta_{ij}^i, \delta_{ij}^j \in [-\Delta, +\Delta] \subset \mathbb{Z}$, which slightly changes the coordinate $p_i$ to a new value $p_i' = p_i + \delta_{ij}^i$ in the interval $[p_i - \Delta, p_i + \Delta]$. The same for $p_j$; its new value $p_j' = p_j + \delta_{ij}^j$ will belong to interval $[p_j - \Delta, p_j + \Delta]$. A new array of pixels is obtained, with randomly distributed positions $(p_i', p_j')$. No specific assumption needs to be made about the distribution of both $s$ and $\delta_{ij}$ at this point.

Figure 1 shows an example of the array of pixels at positions $(p_i, p_j)$ and the new set of pixel with coordinates $(p_i', p_j')$. To each pixel in a round neighbourhood with radius $\rho$ centred at $(p_i', p_j')$, we give the colour tones of the original point, $b(p_i, p_j, c)$. This mixes the colour tones in the image, producing an effect of fleeting colours and lights, as they were passing through an optical inhomogeneous medium. The radius $\rho$ of this round spot is adjusted according to the original image, as a different paintbrush.

All this procedure, that is the determination of the set of pixels and its substitution with a set of spots, is repeated several times to cover the whole image. Each iteration has a different value for the parameter $s$. Figure 2 is an illustration of this iterative procedure. In the following figure (Fig.3), we see the iteration procedure applied to the same input image, with an increased value of the parameter $\Delta$. Note the increase of randomness.

Instead of considering a circular neighbourhood centred at $(p_i', p_j')$, it is possible to use an anisotropic rectangular spot, with two different dimensions, determined comparing the colour tones of pixels. For this comparison, we can use several procedures, as the following one, which is quite simple.

Let us introduce three parameters $\Lambda, \Lambda', \Lambda'' \in \mathbb{N}: \Lambda' < \Lambda < \Lambda''$ and consider the three values: $b_1 = b(p_i, p_j, c)$, $b_2 = b(p_i + \Lambda, p_j, c)$ and $b_3 = b(p_i, p_j + \Lambda, c)$. Then we define two parameters:

$$A = \frac{|b_2 - b_1|}{b_1} ; B = \frac{|b_3 - b_1|}{b_1} \qquad (1)$$

The rectangular neighbourhood of pixel $(p_i', p_j')$ has sides $d', d''$, chosen according to the following scheme:

$$\begin{aligned} A \leq \tau, B \leq \tau &\rightarrow d' = \Lambda, d'' = \Lambda \\ A > \tau, B > \tau &\rightarrow d' = \Lambda', d'' = \Lambda' \\ A \leq \tau, B > \tau &\rightarrow d' = \Lambda'', d'' = \Lambda' \\ A > \tau, B \leq \tau &\rightarrow d' = \Lambda', d'' = \Lambda'' \end{aligned} \qquad (2)$$

where $\tau$ is a fixed threshold value. The result of this rendering, with values $\tau = 0.1, \Lambda = 3, \Lambda' = 2, \Lambda'' = 5$, and working only on the red colour tone, is shown in Figure 4. Note the effect of fleeting lights coming from the surface of water. This is due to the fact that we are using a comparison procedure (1-2) which is done on the original points $(p_i, p_j)$, while the rectangular spots are assigned to the randomly displaced points $(p_i', p_j')$.

In the case that values $b_1' = b(p_i', p_j', c)$, $b_2' = b(p_i'+\Lambda, p_j', c)$ and $b_3' = b(p_i', p_j'+\Lambda, c)$, and parameters

$$A' = \frac{|b_2'-b_1'|}{b_1'} \ ; \ B' = \frac{|b_3'-b_1'|}{b_1'}, \qquad (3)$$

were used instead of $b_1, b_2, b_3, A, B$, in the procedure to determine the spot assigned to position $(p_i', p_j')$, we obtain what is shown in Figure 5. The result is quite different, with a strong reduction of flickering effects.

**4. Discussion**
The previously discussed procedure uses a vertically or horizontally oriented rectangular spot; of course, it would be better to have a rectangular or elliptic spot, oriented according to the peculiar directions of the image texture. In this framework, we are planning to use the Coherence Length Diagrams approach, previously proposed in Ref.18,19.

Other possible renderings could be based on the thresholding method approached. We could process images as follows. After determining the positions with coordinate $(p_i, p_j)$ and $(p_i', p_j')$ as previously described, let us consider a square neighbourhood centred at $(p_i', p_j')$, with a large enough size $\Pi$. Pixels in this neighbourhood are at positions given by $(\xi, \eta)$, where $\xi \subseteq [p_i'-\Pi, p_i'+\Pi]$ and $\eta \subseteq [p_j'-\Pi, p_j'+\Pi]$. Let us define $b' = b(\xi, \eta, c)$, $b = b(p_i, p_j, c)$ and the following parameter:

$$C = \frac{|b'-b|}{b} \qquad (4)$$

and compare the parameter $C$ with a fixed threshold value $\tau'$ for each value of $(\xi, \eta)$. Let us fix the colour tones according to the following condition:

$$b(\xi, \eta, c) = b(p_i, p_j, c), \ when \ C \leq \tau' \qquad (5)$$

This rendering is based on the thresholding method, where the parameters $\Pi$ and $\tau'$ strongly influence the final image. Figure 6 shows the result on the same input image of Figure 4 and 5, which considerably different from the previous one.

Also in this case, as we did to obtain Figure 4, we are using the tones of the original points $(p_i, p_j)$ for the random displaced points $(p_i', p_j')$. The other possibility is to assume $b = b(p_i', p_j', c)$ in Eq.(4). The rendering condition is then written as:

$$b(\xi, \eta, c) = b(p_i', p_j', c), \ when \ C \leq \tau' \qquad (6)$$

The result is given in Fig.7. The effect of fleeting lights turns out to be furthermore reduced.
All described procedures can be organized in a single tool, to have several possibilities of rendering. A package running on Windows NT/2K with a .NET package is under development, working on any BMP or JPG picture.

As we discussed at the beginning, one of the goals in non-photorealistic rendering is to stimulate sensations, with a certain enhancing of patterns in the image scene. If the rendering algorithm turns out to be good for this purpose is, to a certain extent, a rather subjective conclusion.

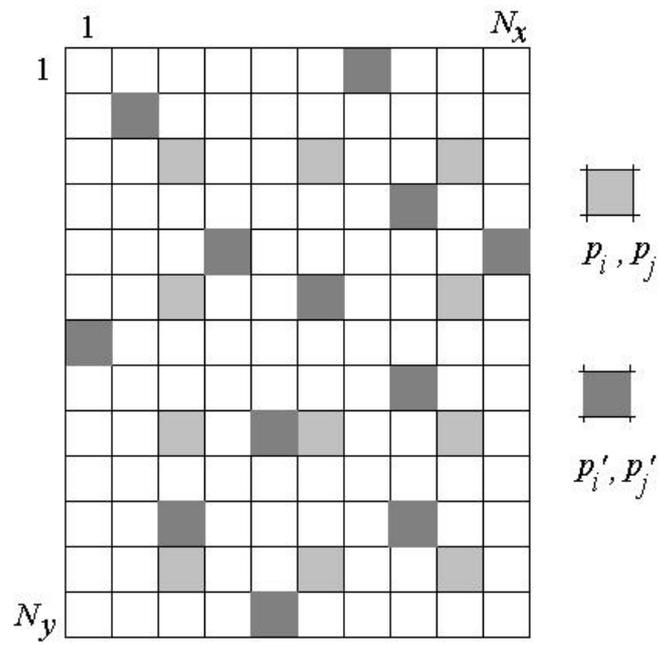

Fig.1 Pixel arrays used for rendering (see text for explanation).

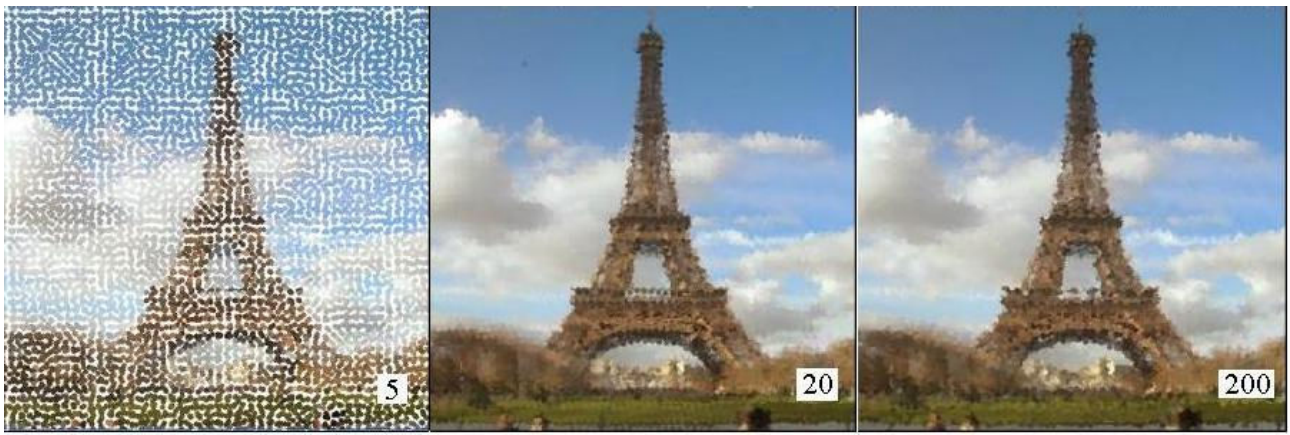

Fig.2 Results obtained after different iteration lengths.

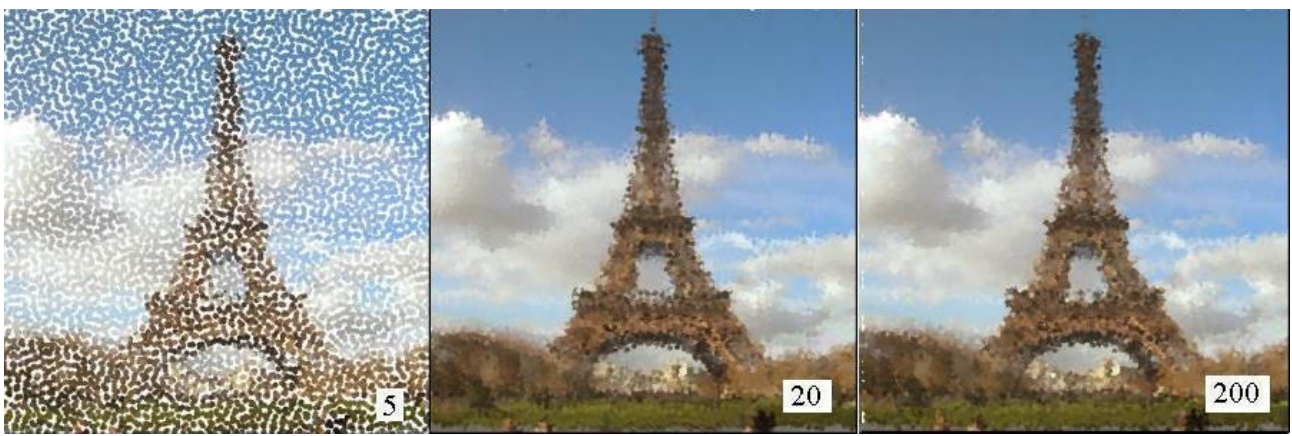

Fig.3 The same as in Fig.2, with a different parameter $\Delta$.

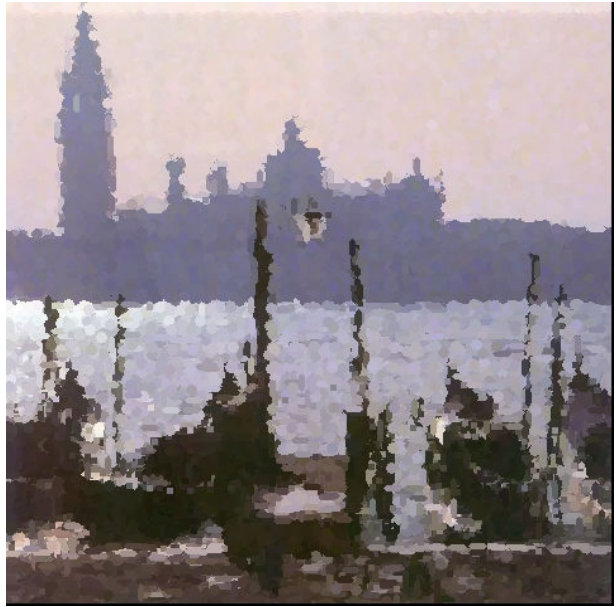

Fig.4 Rendering with rectangular spots (according to Eq.(1-2), see text for explanation).

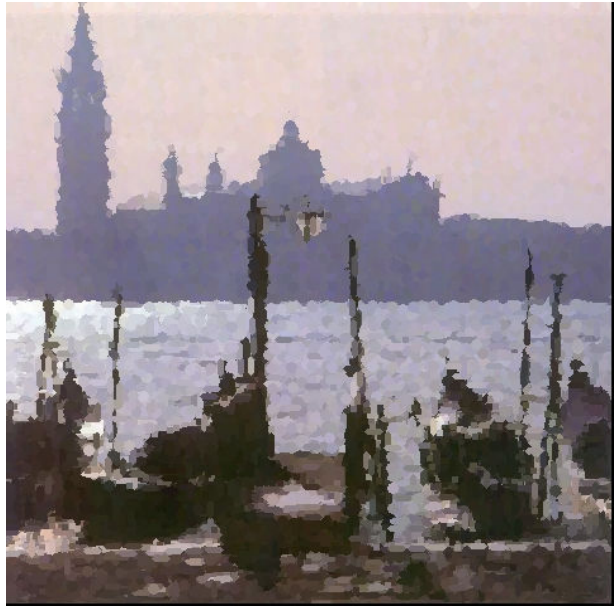

Fig.5 Rendering with rectangular spots (according to Eq.3, see text for explanation).

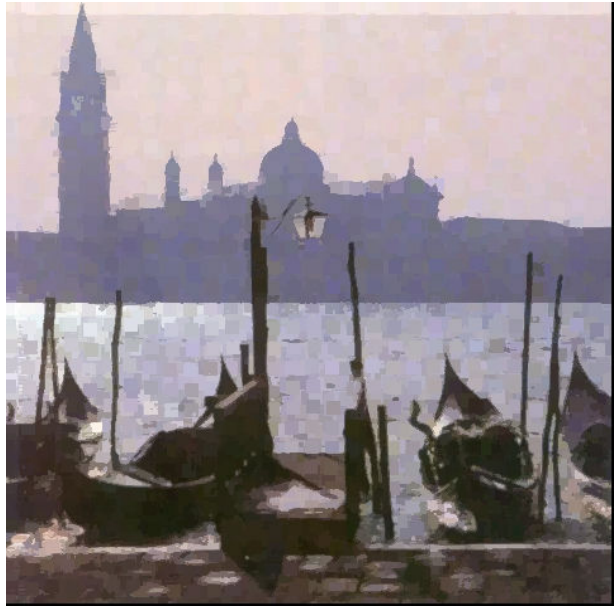

Fig.6 Rendering with a threshold method, according to Eq.4-5.

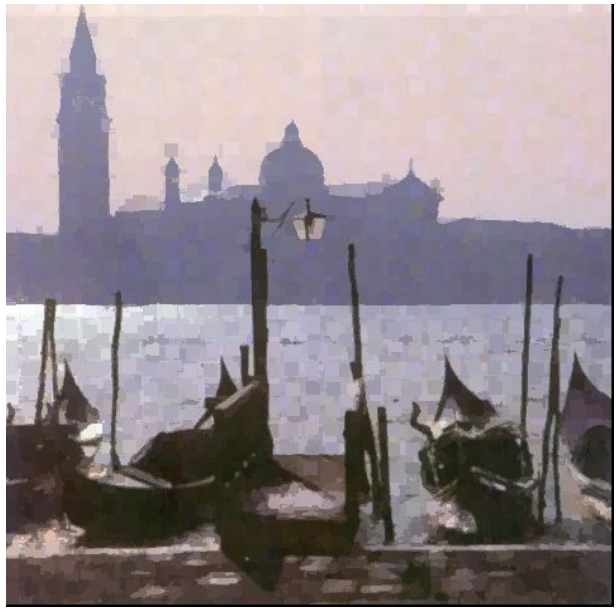

Fig.7 Rendering with a threshold method, according to Eq.6.